\documentclass[a4paper,fleqn]{cas-dc}

\usepackage{natbib}
\setcitestyle{numbers,square}
\usepackage{lineno,rotating,algorithm,algpseudocode}
\usepackage{bm,amssymb,multirow}
\usepackage{amsmath,amsfonts,color}
\usepackage{times}
\usepackage{graphicx}
\usepackage{url}
\usepackage{booktabs}
\usepackage{caption,subcaption}
\usepackage{array}
\usepackage{multicol}
\usepackage{multirow}
\usepackage{url,footnote}
\usepackage{stfloats}
\usepackage{makecell}
\usepackage{microtype,hyperref}
\usepackage{amsmath,amsfonts,amssymb,bm,color}
\usepackage{times}
\usepackage{graphicx}
\usepackage{url}
\usepackage{booktabs}
\usepackage{caption,subcaption,algorithm,algpseudocode,multirow,multicol}
\usepackage{bbding}
\usepackage{pifont}
\usepackage{makecell,tabularx}
\usepackage{subfloat}
\usepackage{array}
\usepackage{longtable}
\usepackage{supertabular}
\usepackage{makecell}
\usepackage{multicol}
\usepackage{multirow}
\def\tsc#1{\csdef{#1}{\textsc{\lowercase{#1}}\xspace}}
\tsc{WGM}
\tsc{QE}
\begin{document}
\let\WriteBookmarks\relax
\def\floatpagepagefraction{1}
\def\textpagefraction{.001}

\shorttitle{Physics Inspired Hybrid Attention for SAR Target Recognition}    

%\shortauthors{<short author list for running head>}  

\title [mode = title]{Physics Inspired Hybrid Attention for SAR Target Recognition}  

%\tnotemark[<tnote number>] 



\author[1]{Zhongling Huang}
%\cormark[1]


% \ead{huangzhongling@nwpu.edu.cn}

%\ead[url]{<URL>}

%\credit{Conceptualization of this study, Methodology, Software}

\affiliation[1]{organization={the BRain and Artificial INtelligence Lab (BRAIN LAB), School of Automation, Northwestern Polytechnical University},
            city={Xi'an},
%          citysep={}, % Uncomment if no comma needed between city and postcode
            postcode={710072}, 
            country={China}}
\affiliation[2]{organization={Anhui Provincial Key Laboratory of Multimodal Cognitive Computation, School of Artificial Intelligence, Anhui University},
            city={Hefei},
%          citysep={}, % Uncomment if no comma needed between city and postcode
            postcode={230601}, 
            country={China}}
\affiliation[3]{organization={Beijing Technology and Business University},
            city={Beijing},
%          citysep={}, % Uncomment if no comma needed between city and postcode
            % postcode={230601}, 
            country={China}}
\author[1]{Chong Wu}
% \cormark[1]
% Footnote of the second author
% Email id of the second author
%\ead{}



\author[1]{Xiwen Yao}
\cormark[1]

\author[2]{Zhicheng Zhao}
\author[3]{Xiankai Huang}
\author[1]{Junwei Han}
% \affiliation[2]{organization={Remote Sensing Technology Institute (IMF), German Aerospace Center (DLR)},
            % city={Wessling},
%          citysep={}, % Uncomment if no comma needed between city and postcode
            % postcode={82234}, 
            % country={Germany}}

\cortext[1]{Corresponding author}

% \fntext[1]{$\dag$: Equal Contribution}


\begin{abstract}
There has been a recent emphasis on integrating physical models and deep neural networks (DNNs) for SAR target recognition, to improve performance and achieve a higher level of physical interpretability. The attributed scattering center (ASC) parameters garnered the most interest, being considered as additional input data or features for fusion in most methods. However, the performance greatly depends on the ASC optimization result, and the fusion strategy is not adaptable to different types of physical information. Meanwhile, the current evaluation scheme is inadequate to assess the model's robustness and generalizability. Thus, we propose a physics inspired hybrid attention (PIHA) mechanism and the once-for-all (OFA) evaluation protocol to address the above issues. PIHA leverages the high-level semantics of physical information to activate and guide the feature group aware of local semantics of target, so as to re-weight the feature importance based on knowledge prior. It is flexible and generally applicable to various physical models, and can be integrated into arbitrary DNNs without modifying the original architecture. The experiments involve a rigorous assessment using the proposed OFA, which entails training and validating a model on either sufficient or limited data and evaluating on multiple test sets with different data distributions. Our method outperforms other state-of-the-art approaches in 12 test scenarios with same ASC parameters. Moreover, we analyze the working mechanism of PIHA and evaluate various PIHA enabled DNNs. The experiments also show PIHA is effective for different physical information. The source code together with the adopted physical information is available at \textcolor[RGB]{22,12,202}{\url{https://github.com/XAI4SAR/PIHA}}.

\end{abstract}


%\item 
%\item 
%\item 

% Keywords

\maketitle

% Main text
\section{Introduction}
\begin{figure}[!htbp]
\centering
\includegraphics[width=8cm]{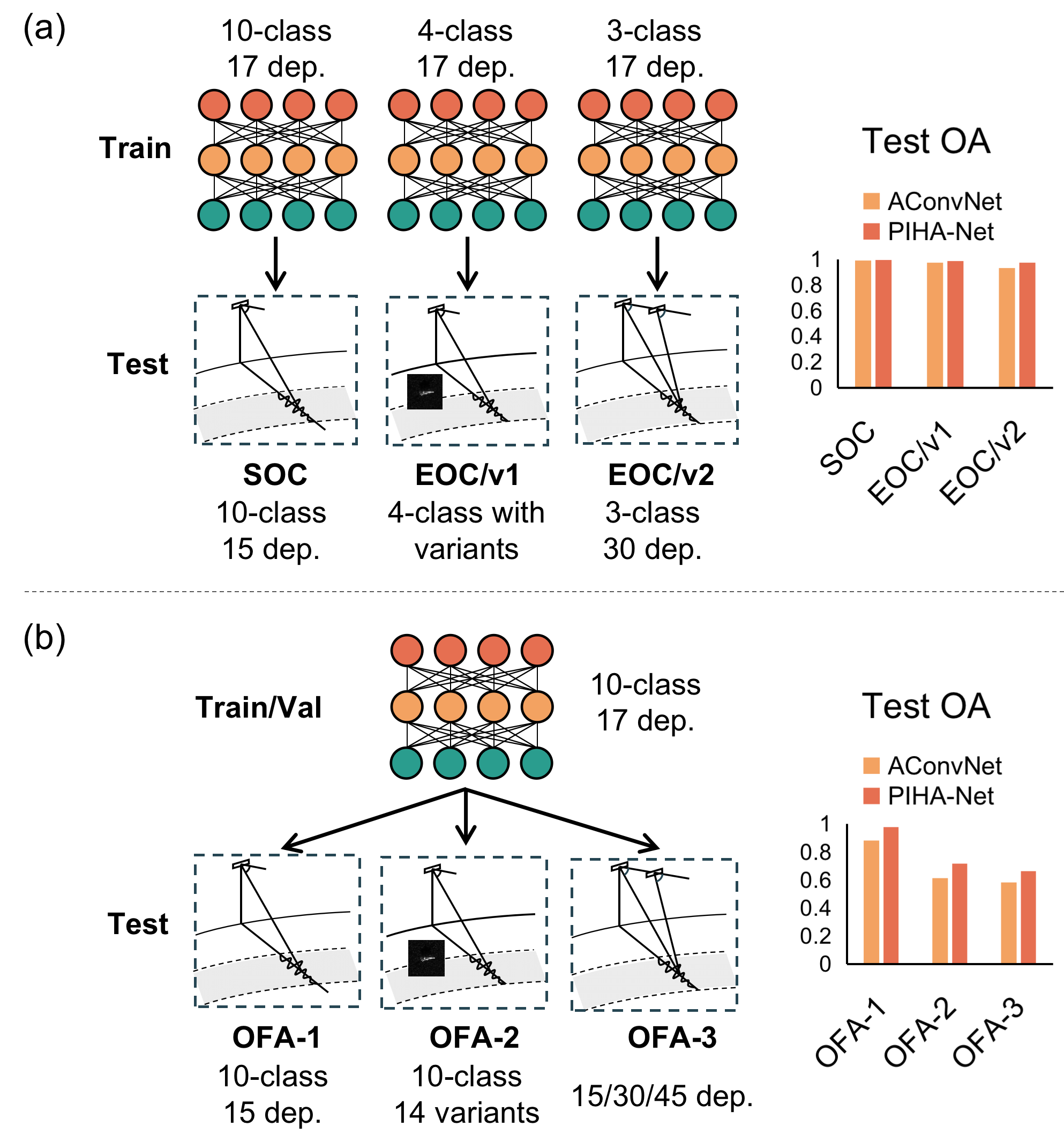}

\caption{The overall accuracy (OA) of both A-ConvNet and the proposed PIHA-Net exhibit very high performance when evaluated using the conventional SOC/EOC setting. However, there is a notable disparity in their results when assessed using the proposed Once-For-All (OFA) evaluation protocol, which was conducted on the MSTAR dataset. The proposed OFA evaluation is more challenging to assess the generalization ability and robustness of a method.}
\label{fig:intro}
\end{figure}

Synthetic Aperture Radar (SAR) target recognition has been facilitated by deep learning techniques in recent years. The deep neural networks, for example, can achieve the recognition accuracy of over 99\% on the Moving and Stationary Target Acquisition and Recognition (MSTAR) dataset \cite{mstar} under standard operation condition (SOC) and extended operation conditions (EOC) \cite{7460942}, as shown in Fig. \ref{fig:intro} (a). These success stories are grounded in the data-driven nature of the approaches and the straightforward experimental settings.

In addition to attaining greater accuracy for SAR target recognition with more complex deep learning models, there is an increasing demand for the robustness and generalization capability in open environment \cite{zhou2022open}, as well as the interpretability of model design, the consistency of scientific theories, and the awareness of domain knowledge \cite{grsm2023,pxdl2022}. These requirements have prompted an increase in research efforts aiming at enhancing deep learning models through the integration of prior knowledge into the learning process. It is referred to informed machine learning \cite{9429985}, or theory-guided data science (TGDS) \cite{karpatneTheoryguidedDataScience2017}.

We preliminarily probe into the integration paradigms of deep neural networks and the physical information of SAR for SAR image interpretation \cite{pxdl2022,grsm2023}. One of them is the cascade and fusion mode, and another is physics inspired design of model architecture. In the field of SAR target recognition, the physical characteristics of SAR targets, especially the physical parameters of scattering centers, have garnered significant attention as an important branch of domain knowledge \cite{zhangFECFeatureFusion2021,10068265}. Many recent studies successfully integrate the physical scattering model of SAR into the deep neural networks by considering the ASC parameters of target as an additional data or feature for fusion \cite{liuMMFFMultimanifoldFeature2021a,DSN2020}. However, the model performance is highly influenced by the fixed input physical features. Although the physical features are robust and interpretable, they may be inadaptable for the current task. Besides, the fusion strategy cannot be popularized to adapt various physical information.

To address this issue, we propose to design a knowledge inspired module for SAR target recognition that leverages the physical information in a more flexible way. It can be adaptable to diverse physical information, transferable to arbitrary DNNs, and less influenced by the specific physical feature values.

\begin{figure}[!htbp]
\centering
\includegraphics[width=8cm]{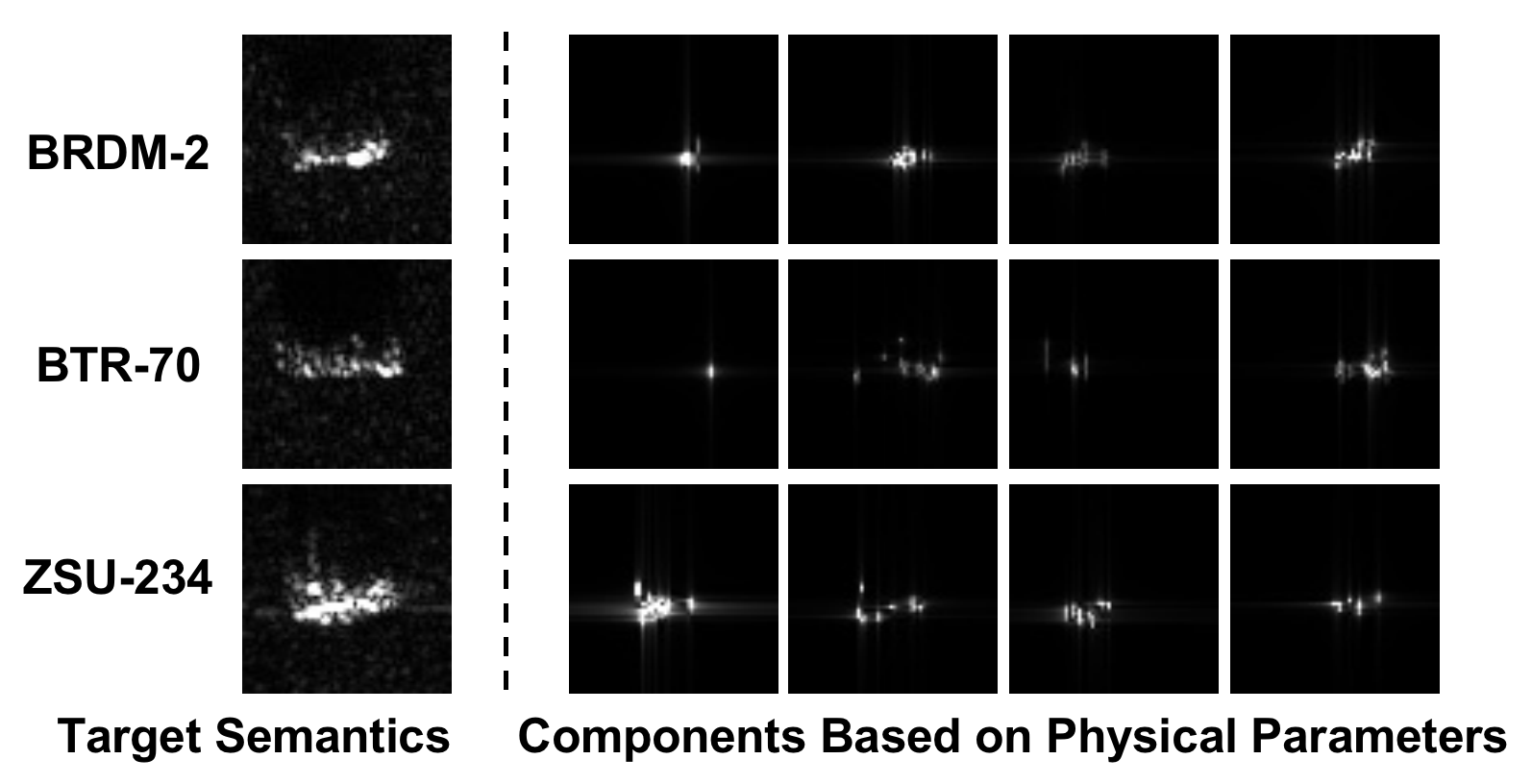}
\caption{On the basis of the physical model, such as the attributed scattering center (ASC), SAR target can be construed as multiple components with local semantics. It provides strong knowledge prior for target recognition.}
\label{fig:intro2}
\end{figure}

As shown in Fig. \ref{fig:intro2}, the prominent scattering centers are clustered based on ASC parameters. Each cluster represents a physically explainable component of the target that describes the local semantic, providing rich knowledge prior in recognizing SAR target. Instead of considering the physical parameters as input data or feature, the high-level semantics of physical information are leveraged in this work. We propose the physics inspired hybrid attention (PIHA) mechanism that utilizes the target components shown in Fig. \ref{fig:intro2} to activate different groups of features and then guide the feature importance re-weighting, making it aware of local semantics. It is flexible to leverage the physical information of SAR target, with the valuable prior resulting in the more contrasting re-weighting of features to influence the outcome. In comparison, the inferior prior will have little impact on feature re-weighting, which prevents performance degradation. PIHA benefits from the flexibility of neural network fitting and the scientificity of the semantic-aware physical information.

% for SAR targets that leverages the physical priors shown in Fig. \ref{fig:intro2}. It benefits from the flexibility of squeeze and excitation (SE) data-driven channel attention and the scientificity of the semantic aware physical information, and can be inserted into various deep architectures.

In order to evaluate the generalization and robustness of PIHA more effectively, we propose a rigorous evaluation protocol once-for-all (OFA) for the popular MSTAR dataset, i.e., once a deep learning model has been trained and validated on known data, it is evaluated on multiple blind test sets with diverse data distributions. It is more challenging than conventional SOC and EOC evaluation protocols, as shown in Fig. \ref{fig:intro} (b). Notably, many related works have used differing optimization results of the physical model (ASC) to integrate with deep learning for SAR target recognition, which makes it difficult to compare the results with each other fairly. Therefore, we carry out a comparative study based on unified physical information and make it publicly available. In addition, we examine in depth the working mechanism of hybrid attention and summarize the benefits and drawbacks of physical constraints in hybrid modeling. We think that the conclusion can illuminate the design of a deep model that incorporates SAR domain knowledge.

The contributions of our work are summarized as follows:

\begin{enumerate}
    % \item A novel physics inspired hybrid attention (PIHA) mechanism is proposed for SAR target recognition, where the advantages of data-driven  and physical information are combined. It can be embedded into different deep architectures to improve the performance.
    \item On the basis of knowledge-guided model architecture design, a novel physics-inspired hybrid attention (PIHA) mechanism is proposed for SAR target recognition, in which the semantic prior of physical information is adaptively incorporated with the attention mechanism. It is flexible for different types of physical information and can be incorporated into various deep architectures to enhance performance.

    \item In PIHA mechanism, a novel physics-driven attention module named PASE is proposed to utilize the physical information of SAR target for activation and feature re-weighting. The high-level semantics of physical information is leveraged in PASE to ensure the flexibility and generalizability for various physical information. 
    
    \item In the experiments, we propose the once-for-all (OFA) evaluation protocol to thoroughly assess the algorithm, demonstrating the robustness and generalization capabilities more effectively. Furthermore, this study delves into the comprehensive examination of the effects of data-driven and physics-driven attentions, offering valuable insights that can serve as a source of inspiration for design concepts.
    
    \item The physical information of SAR targets used in this study together with the source code are open to public, which ensures the reproducibility of our work and facilitates a fair comparison of the results with other methodologies.
\end{enumerate}

In Section \ref{sec:relwork}, we review the related work of SAR target recognition approaches that integrate the physical model and deep neural networks for hybrid modeling, as well as the attention mechanism in computer vision filed. The proposed PIHA module is introduced in Section \ref{sec:method}. Section \ref{sec:exp} presents the experiments and discussions and the conclusion is given in Section \ref{sec:conclusion}.

\section{Related Work}
\label{sec:relwork}

\subsection{SAR Target Recognition with Hybrid Modeling}

Compared to optical images, SAR images possess a number of distinct characteristics, which can be described by some physical models such as attributed scattering center or the specific data format like phase information in complex data. We define the approaches integrating the physical model or physical information of SAR with deep neural networks as hybrid modeling \cite{grsm2023}. 

The most prevalent and efficient method is feature fusion with a cascaded physical model in a neural network. The principle concern is the representation of the physical model. For example, based on ASC model, the previous work applied the bag of words (BoW) \cite{zhangFECFeatureFusion2021}, point cloud \cite{9773339}, as well as the reconstruction part image \cite{9884621}, to represent the ASC parameters as efficient features that can be easily fused with image features. The other concern is the fusion approach, including early fusion (data), mid-level fusion (feature), and late fusion (decision). For instance, the feature fusion based on discriminant correlation analysis (DCA) \cite{zhangFECFeatureFusion2021}, cross attention \cite{10068265}, concatenate or addition of global image feature and part image features \cite{liMultiscaleCNNBased2021}, adaptive weighting fusion \cite{9751685}, multi-manifold fusion of amplitude and phase features \cite{liuMMFFMultimanifoldFeature2021a}, and fusion in decision level \cite{9556630}.

% Some approaches \cite{liuMMFFMultimanifoldFeature2021a, 9780199} take advantage of information of complex data to enrich the feature.

% The extracted ASC parameters can be processed by many ways such as bag of visual words (BOVW) \cite{zhangFECFeatureFusion2021} , point cloud network \cite{9773339} , Inverse Fast Fourier Transform(IFFT) \cite{9365697} and ASC part reconstruction\cite{9884621} to fuse with visual feature. PAN \cite{10068265} uses cross attention between ASC part feature and global feature to contribute better integration. 
% Besides the above properties, strong scattering points \cite{10137878} are extracted in SFSA(SSP), and then the feature is fused with visual feature. IFLS \cite{9751685} processes the feature of target area and shadow area and adopts an adaptive weights during the fusion. Apart from the feature level fusion, \cite{9556630} adopts decision level fusion strategy where final result is decided by prediction results of global feature and ASC part feature.
%即使这些方法效果已经很好，但是仍然存在一些缺点，可解释性较差，简单融合物理信息难以实现物理信息地高效利用。

The other hybrid modeling paradigm is physics inspired or physics guided model design for SAR target recognition \cite{pxdl2022}. Karpatne et al. \cite{karpatneTheoryguidedDataScience2017} proposed the theory-guided data science (TGDS) paradigm, including the theory-guided design of model architecture in which scientific knowledge is used to influence the architecture of data science model. In SAR target recognition field, Liu et al. \cite{liuEFTLComplexConvolutional2021} proposed to initialize the model weights based on ASC parameters. Feng et al. \cite{9896887} designed a physics guided neural network to embed the scattering center knowledge into feature representation motivated by \cite{physically2022}.

The success of feature fusion-based hybrid modeling approaches is primarily dependent on the high representation capacity of physical features and the efficacy of the fusion strategy. It is less adaptable to various physical characteristics. For algorithm development, the physics-guided model design is more interpretable and transparent. Both of them can achieve more robust and generalized model. 

% While these approaches have achieved promising results, there still exists some disadvantage: 1. Simple fusion strategy results in poor interpretability. 2. SAR properties concentrate on representing physical meaning rather than classification which leads to inefficient use of SAR properties. TGDS \cite{karpatneTheoryguidedDataScience2017} summarizes many research themes, such as parameter initialization and data augmentation, to integrate scientific knowledge and data science efficiently.  EFTL \cite{liuEFTLComplexConvolutional2021}   \cite{8651446} purposes a data augmentation method by masking the scattering center of target to enhance robustness of model.

% Different from the above method, We follow the TGDS \cite{karpatneTheoryguidedDataScience2017} to design a theory-guided network. By clustering the parameters of ASC, we divide the target into components which will be reweighted by channel attention and the visual feature will be supplemented with physical feature to enhance robustness. Thus, a scientific and interpretable method is achieved.

%不同于以上方法，我们遵循TGDS的思路对网络模型的结构进行设计，我们基于提取ASC参数并基于该参数将目标分成多个部件，通过对部件间进行使用注意力进行权重分配，我们实现了科学的可解释的深度学习。

\begin{figure*}[!htbp]
\centering
\includegraphics[width=15cm]{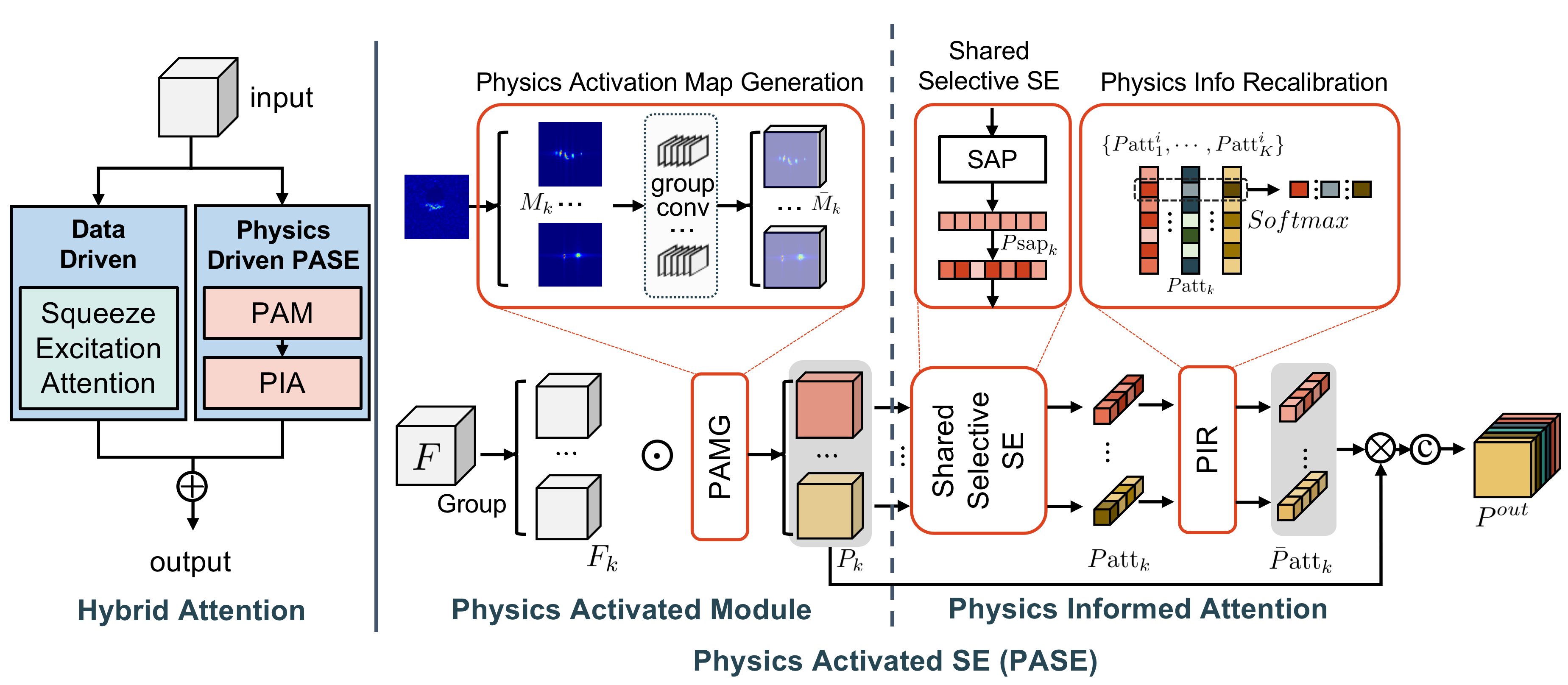}
\caption{The proposed physics-inspired hybrid attention (PIHA) is a combination of data-driven and physics-driven attention PASE. The physics-activated squeeze and excitation module (PASE) is composed of physics activated module (PAM) and physics-informed attention (PIA).}
\label{fig:piha}
\end{figure*}

\subsection{Attention Mechanism in Deep Learning}
%深度学习中的注意力机制（Attention Mechanism）是一种模仿人类视觉和认知系统的方法，它允许神经网络在处理输入数据时集中注意力于相关的部分。\cite{}
%spatial and channel attention, principles, cross-attention/self-attention/,...
The attention mechanism attempts to imitate the human visual ability and cognitive system in order to focus on the significant information of the image. SE \cite{8578843} proposed channel attention to reweight the feature in each channel. ECA-Net \cite{9156697} adopted an 1D convolution to decrease the complexity of generating channel attention. Further, SK-Net \cite{8954149} and EPSANet \cite{Zhang_2022_ACCV} split feature along channel dimension and process feature with kernel of different size to acquire multiscale channel attention. A2-Nets \cite{chen20182} introduced a novel function to capture the global relation of spatial information. CBAM \cite{woo2018cbam} and BAM \cite{Park2018BAMBA} enriched the attention map by combining the channel attention and spatial attention. DANet \cite{8953974} applied the self-attention mechanism on channel attention and spatial attention to model long-range dependency. SANet \cite{zhang2021sa} split the feature along the channel dimension and adopt channel attention and spatial attention on each group.

The above mentioned methods aim to optimize the structure of channel and spatial attention to achieve better performance. The calculation of attention is solely dependent on the given feature map, and we summarize them as data-driven attention. Another feasible way is combining attention with prior knowledge. Hao et al. \cite{hao2017end} applied a pre-trained embedding matrix, which contains the information of a large knowledge base, to build the cross attention model in question answering task. He et al. \cite{he2022named} adopted TransE \cite{bordes2013translating} to embed the marine knowledge graph into a self-attention module in named entity recognition task. In medical image classification, PKA2-Net \cite{fu2023pka} generated a series of templates on principle of energy decay to obtain the attention feature. In heart disease prediction, MINA \cite{hong2019mina} combined the prior medical knowledge \cite{kashani2005significance, yanowitz2012introduction} to compute attention weights.

%上述方法要么要么。我们
% The reviewed attention methods either focus on the structure optimization or the integration of prior knowledge.  In order to take advantage of benefits of two type attention, PASE is proposed, which is designed on principle of physics model and leverage various types of physical knowledge of SAR targets.

The data-driven attention is flexible and adaptive, while the knowledge integrated attention is aware of prior information. In order to take the advantage of both, we propose the hybrid attention PIHA in this paper.

\section{Method}
\label{sec:method}

%PIHA结构如图x所示，它由数据驱动支路以及物理驱动支路组成，其中物理驱动支路由PAM和PIA组成。在本部分，我们首先介绍数据驱动支路并解释其存在的意义。然后我们将介绍PASE模块中的PAM和PIA模块，PAM基于ASC模型的对视觉特征进行物理激活，PIA对物理激活特征进一步细化调整。%

In this section, we first introduce the overall structure and design idea of the proposed PIHA. Then, the details of the data-driven and physics-driven block, named physics activated squeeze and excitation (PASE) module, is illustrated. Finally, we present how to design a PIHA enabled deep neural network.
% effect of two block and briefly summarize the structure of data driven block. Next, combining the attribute scattering center (ASC) model, the physics activated squeeze and excitation module (PASE), which composed of physics activated module (PAM) and physics informed attention (PIA), will be introduced in detail.

% \subsection{Effect of data driven block and physics driven block}
\subsection{PIHA Overview}

The physical information of SAR target is robust but less flexible and may be not exactly to the benefit of recognition. Different from the most hybrid modeling methods that consider the physical information as an input data or feature for fusion, the proposed method leverages the high-level semantics provided by physical information to re-weight features. PIHA is a combination of data-driven and physics-driven blocks, as shown in Fig. \ref{fig:piha}, that regulates the impact of semantic prior adaptively. The physics-driven block, denoted as physics activated squeeze and excitation (PASE), aims to activate and re-weight the feature groups by means of knowledge prior to make them aware of local semantics. The valuable prior results in contrasting feature re-weighting that highlight important features to influence the result. The data-driven SE attention dominates when knowledge prior is inferior. The design ensures the positive impact of physical information and improves the flexibility of applying physical information. Besides, PIHA is a general module that can be inserted to any level of DNNs and adapts different physical models.

% PIHA, which blends the conventional SE and physics activated SE, leverages the flexibility and adaptivity of neural network and the knowledge prior of physical information to obtain better generalization ability and robustness.

% However, despite of the strong robustness of physical feature, it is not recommended to merely apply physical feature to classification task because physical feature focus on physical meaning representation rather than class distinguishing.

Let $F\in\mathbb{R}^{C\times H \times W}$ denotes the input feature, where $C,H,W$ represent channel, height, weight respectively. The data-driven attention implemented by a SE block \cite{8578843} aims to re-weight the feature map $F$ based on the global visual information, obtaining $D^{out}$. On the other hand, the physics-driven attention, denoted as PASE module, respectively enhances the feature $F$ based on the local semantics of physical information and outputs $P^{out}$. The eventual output of PIHA is denoted as:
\begin{equation}
   F_{\mathrm{PIHA}} = D^{out} + P^{out}, 
\end{equation}
as the fusion of data-driven and physics-driven outputs.

\subsection{Data-Driven SE}

The data driven SE block consists of squeeze and excitation for decoding the global spatial information and re-calibrating weights of each channel, which can be referred to literature \cite{8578843}. First, global average pooling is adopted to squeeze the spatial information:
\begin{equation}
% \label{deqn_ex1}
D_c^{sq}=\frac{1}{{H\times W}}\sum_{i=1}^H\sum_{j=1}^WF_c(i,j)
\end{equation}
where $F_c$ denotes the c-th channel of input feature. Then, a bottleneck with two fully-connected layers, i.e. a dimensionality-reduction layer and a dimensionality-increasing layer, are adopted to generate channel attention $D^{attn}\in\mathbb{R}^{C\times 1 \times 1}$: 
\begin{equation}
\label{equ:se}
D^{attn}=\sigma (W_1\delta (W_0(D^{sq})))
\end{equation}
where $W_1\in\mathbb{R}^{\frac{C}{r}\times C}$ and $W_0\in\mathbb{R}^{C\times \frac{C}{r}}$. $\delta$ represents Recitified Linear Unit (ReLU) \cite{10.5555/3104322.3104425} operation and $\sigma$ is excitation function where Sigmoid is usually used. Finally, the input feature is recalibrated by the channel attention:
\begin{equation}
% \label{deqn_ex2}
D^{out}=D^{attn}\otimes F
\end{equation}
where $\otimes$ refers to the channel-wise multiplication.

% \begin{equation}
% \label{deqn_ex1}
% F_{\mathrm{PIHA}} = D^{out} + P^{out}
% \end{equation}
\begin{table}  
\centering
\caption{The combination of $\alpha$ and $L$ defines canonical types of scattering centers.}
\label{tab:asc}
\begin{tabular}{c c c}
\toprule
    Scattering geometry &$\alpha$& L \\
    \hline
    Trihedral&1&$L=0$\\
    Top Hat&0.5&$L=0$\\
    Corner diffraction&-1&$L=0$\\
    Sphere&0&$L=0$\\
    Edge broadside&0&$L>0$\\
    Edge diffraction&-0.5&$L>0$\\
    Dihedral&1&$L>0$\\
    Cylinder&0.5&$L>0$\\
\bottomrule
\end{tabular}
\end{table}

\begin{figure}[!t]
\centering
\includegraphics[width=3.25in]{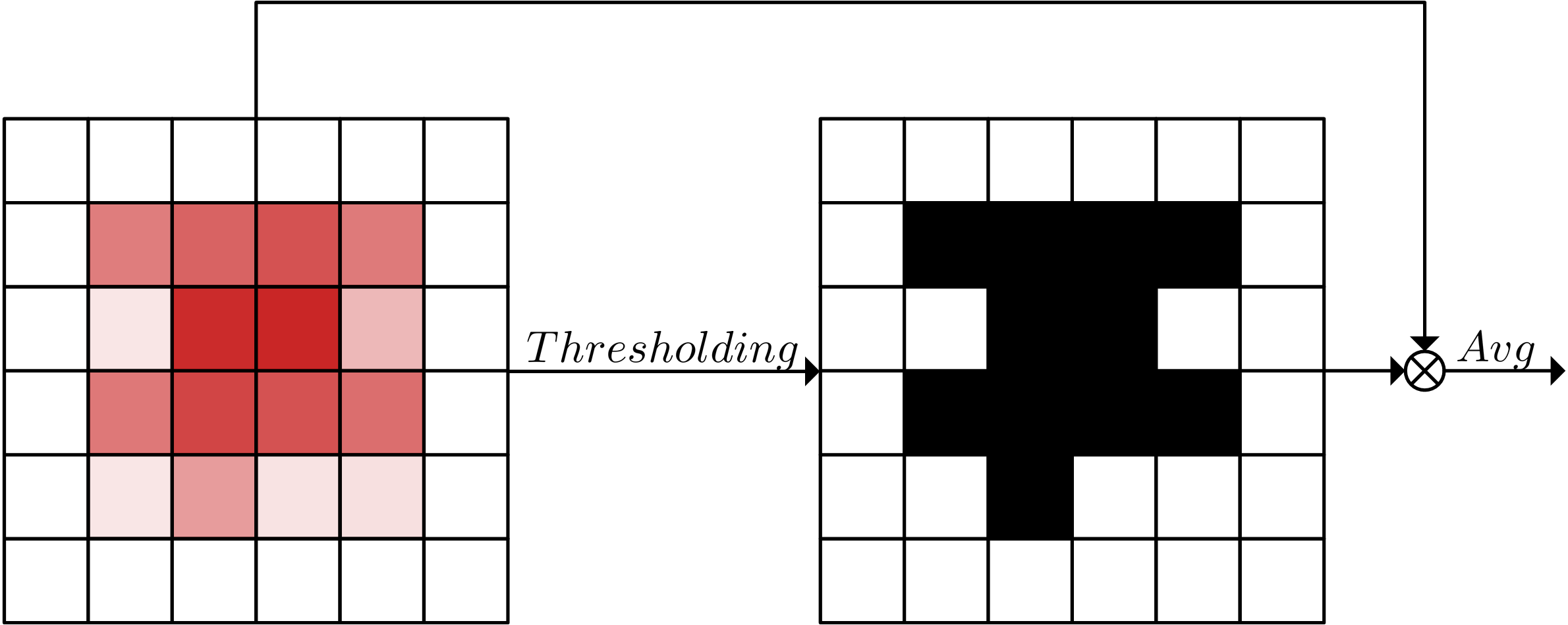}
\caption{Illustration of selective average pooling.}
\label{fig:savg}
\end{figure}

\subsection{Physics-Driven PASE}
% 介绍部分，介绍该模块的动机
% 基于数据驱动的识别算法在数据分布相近的情况下表现优异，然而实际情况中，SAR图像成像效果随俯仰角，成像算法等多种因素的影响。这种情况下基于数据驱动的方法表现不佳，因此我们提出了PASE，通过特征融合对数据驱动的视觉特征进行补充，使得算法在具备强大数据拟合能力的同时，面对不同数据分布的SAR图像也能有很好的表现。PASE由PAM和PIA两部分组成，其中PAM基于物理信息对输入特征空间上的激活，物理激活特征通过PIA对不同物理信息产生联系，并对物理特征进行通道上的权重重分配。
% 输入 input feature
The physics-driven PASE consists of the physics activated module (PAM) and the physics informed attention (PIA).
% Data-driven SAR ATR methods achieve excellent performance when depending on powerful data fitting ability. However, result of SAR imaging is affected by many factors such as depression angle and imaging algorithm and data driven SAR ATR methods is underperforming when data distribution varies considerably. 

As shown in Fig. \ref{fig:intro2}, the target components based on ASC physical model provide rich local semantics that should be valuable for recognizing different targets. Regarding it as a robust prior, PASE is proposed by introducing the physical information to architecture design, achieving physics-aware attentions based on local semantics of target components. Hence, the feature map is re-weighted with channel-wise physics specificity to enhance the generalization capability and robustness.

The proposed PASE is cascaded by PAM and PIA, where PAM activates the spatial information of each sub-group features based on the target component obtained with physical information, and PIA models the channel importance of each sub-group features with local semantics and enhances the interaction of different local semantics using the physical information re-calibration (PIR) cell.

%PASE is proposed to address this issue by complementing visual feature with physics feature so that 这句话放这里不太好，因为这里讲的是物理特征，不是overall  
% 1) \emph{Attribute Scattering Center (ASC) Model}
\subsubsection{Knowledge based Target Components}

In this work, we mainly apply the ASC physical model to generate the target components with physical meanings. Based on the high-frequency approximations, the radar backscattering of a target can be approximated by a set of points. From Gerry's theory \cite{gerry1999parametric}, each scattering center can be represented by seven parameters which is named attributed scattering center. Thus, the target backscattering is parameterized on the basis of geometrical theory of diffraction (GTD) as:
% ASC模型是基于GTD原理而构造的物理模型，该模型被广泛应用于SAR识别算法中，该模型假设SAR的回波可以由多个散射点的回波相应直接叠加
% \begin{equation}
% \label{deqn_ex1}
% E(f,\varphi;\Theta)=\sum^q_{i=1}E_i(f,\varphi;\Theta)
% \end{equation}
\begin{equation}
\begin{aligned}
\label{deqn_ex1}
E(f,\varphi;\Theta)=\sum^q_{i=1}&A_i\cdot(j\frac{f}{f_c})^{\alpha_i}\cdot\exp(2\pi\gamma_i f\sin\varphi) \\
&\cdot \exp[-j4\pi\frac{f}{c}(x_i\cos\varphi+y_i\sin\varphi)]\\
&\cdot sinc(2\pi\frac{L_i}{c}f\sin(\varphi-\bar{\varphi}_i))\\
\end{aligned}
\end{equation}
where $f$ and $\varphi$ denote the frequency and aspect angle, while $f_c$, $c$ and $q$ denote the center frequency, propagation velocity of electromagnetic wave, and the number of scattering centers, respectively. $\Theta$ is the physical parameter set of scattering centers containing $(A,x,y,\alpha,L,\bar{\varphi}, \gamma)$, that are, amplitude $A$ of echos, relative position $(x,y)$, frequency dependence $\alpha$, length $L$, orientation $\bar{\varphi}$, and aspect dependency $\gamma$. The combination of $\alpha$ and $L$ decides the type of scattering centers, as shown in Table \ref{tab:asc}.
% represents the aspect dependency of localized scattering centers. $\alpha$ represents frequency dependence which decide type of scattering centers. Details are shown in table 1.
% The backscattering of each scattering center can be modeled on the basis of geometrical theory of diffraction (GTD) as below:
% where $f_c$ is center frequency and c is propagation velocity of electromagnetic wave.  
%后面讲一下基本的优化问题以及使用了OMP

The estimated physical parameter set $\hat{\Theta}$ is obtained by solving the following optimization problem:

% \begin{equation}
% 	min \enspace q, s.t.\Vert S(f,\varphi)-\sum_{i=1}^q E_i(f,\varphi;\Theta_i)\Vert_2\leq\sigma
% \end{equation}
\begin{equation}
	\hat{\Theta} = \arg\min_{\Theta} \Vert S(f,\varphi)-E(f,\varphi;\Theta) \Vert_2
\end{equation}
where $S(f,\varphi)$ is the signal in frequency domain. In our work, the optimization problem is solved by orthogonal matching pursuit (OMP) algorithm \cite{7862730}.

%ASC参数通常难以被直接应用于分类任务，因为物理参数存在其物理意义，这与视觉特征通常存在巨大差异，因此通常使用统计算法对参数进行处理。
%为了将物理信息更好应用于分类，我们参考了【】，基于BOVW方法获取简单高效的物理特征，K-means算法被引入对ASC散射点进行分类，距离计算如下
%为了获取高效的物理特征，我们选取部分物理参数作为k-means聚类依据，并选取n为4，聚类结果如图所示
Then, the target components are generated based on the estimated physical parameter $\hat{\Theta}$. Following the statistical method in literature \cite{zhangFECFeatureFusion2021}, the bag of visual words (BOVW) representation is adopted to process the ASC feature, followed by K-means clustering to obtain K groups of parameter set, denoted as $\{\Theta_i^{(1)}\}_{i=1}^{N_1}, ..., \{\Theta_i^{(K)}\}_{i=1}^{N_K}$, where $N_k$ denotes the number of the $k$th cluster. Finally, the physical parameters in each group is reconstructed to SAR image domain to obtain K images with target components, denoted as $M_1, ..., M_K$.

In the experiments, we compare the results of applying different parameters in $\Theta$ for clustering, including $\{x,y,L,\bar{\varphi},\alpha,A\}$, $\{x,y,A\}$, and $\{x,y\}$. Apart from ASC model which requires the complex-valued SAR image, we can also adopt other approaches to generate the target components, as shown in Fig. \ref{fig:exp_phyinfo}, which will be further discussed in the experiments.

% Generally, it is tough to explicitly apply ASC parameters to classification tasks because the physical meaning of parameters exist significant difference between visual feature. Thus, statistical methods are usually applied in advance. In our proposed method, bag of visual words (BOVW) is adopted to process ASC parameters.
% In the process of BOVW, K-Means is introduced to divided samples into $k$ parts according to the input feature. We assume each single scattering center as a sample and the corresponding parameters as input feature. We select $\bar{\theta}$, a subset of $\theta$ which is effective in classification, as input feature. The Euclidean distance of $\bar{\theta}$ is:
% \begin{flalign}
% \label{deqn_ex1}
%  &&d=\Vert\bar{\theta}_i - \bar{\theta}_j \Vert_2 \qquad \qquad \bar{\theta}_i,\bar{\theta}_j \in \Theta
% \end{flalign}
% d=\sqrt[d]{|x_1-y_1|^q+|x_2-y_2|^q+\cdots+|x_n-y_n|^d} 
% where $\Theta$ denotes set of all the scattering centers. Number of centers k is set to 4 and the ASC based part maps are shown in Fig. 5.

%The imaging process of SAR emit electromagnetic waves which obeys the GTD and receive the 

% \begin{figure}[!t]
% \centering
% \includegraphics[width=3.25in]{cluster.png}
% \caption{Clustering result of BOVW.}
% \label{fig1}
% \end{figure}

% 2) \emph{Physics Activation Module}

\subsubsection{Physics Activation Module}
% 激活方法，首先从对物理特征的处理上说，先说方法，再说原因，然后从输入特征说
%在SAR图像的分类中，虽然SAR图像已经具有很高的分辨率，但是相对于高分辨的光学图像，SAR目标显得十分模糊，只能获取简单的轮廓信息，难以从目标上获取目标的结构信息。PAM被提出解决该问题，由于ASC模型将模型分为多个散射点，相比于整个目标散射互相影响，ASC将散射信息解耦因此散射点相对独立，并且在上一步的统计模型中，性质相似的点被聚类的一起，因此相同的part可以代表组成目标的部件。
% 为了对输入特征进行物理性激活，PAM被提出。在上一步中物理激活图P R被得到。首先，我们对Y进行单通道的分组卷积，这样做第一是为了物理特征经过卷积层通道加深语义更加丰富，同时可以对物理激活图进行大小的调整，第二不同part经过不同的卷积，保证了part之间的独立性。第二步，输入特征X R经过split变成R,

% 
% With the development of SAR, SAR images are getting higher resolution.（没用的话）
% However, SAR images are still blurred comparing to sharp optical images and the inner structure of target is still confusing.（blur跟分辨率没关系，前后句子没有逻辑性）
% Different from backscattering of target interacting with each other, ASC decouples the backscattering of targets as sum of multiple independent scattering centers. After the process of statistic model, scattering centers with similar attribute are clustered resulting in different component of target.

In order to attain the features aware of different local semantics, the input feature $F$ is firstly split into K groups ($F_1, ..., F_K$) and then activated respectively on the basis of the obtained target components $M_k$. As shown in Fig. \ref{fig:piha}, the physics activation map $\bar{M}_k$ is generated by a group of convolutions on each $M_k$, denoted as:
% In the process of PAM, we assume $M\in\mathbb{R}^{k\times \bar{H} \times \bar{W}}$ is original ASC based part map with original image width $\bar{W}$ and height $\bar{H}$. First of all, the i-th part of ASC based activation map $\bar{M}\in\mathbb{R}^{k\times\frac{C}{k}\times H \times W}$ is obtained by a group convolution layer:
% $\bar{M}\in\mathbb{R}^{k\times\frac{C}{k}\times H \times W}$
\begin{equation}
    \bar{M}_k=Conv_k(M_k)
\end{equation}
% \label{deqn_ex1}
% &&\bar{M}_i=Conv_i(M_i)    %\qquad  i=1 \cdots K
% \end{flalign}
where $Conv_k$ is the k-th group convolution layer with a kernel number of $\frac{C}{K}$. In this way, the processed physics activation map $\bar{M}_k$ is with the size of $\frac{C}{K} \times H \times W$, as same as $F_k$. The group convolution operation preserves the local semantics of each $M_k$ while simultaneously refining the feature in each channel within a group. Then, the physics activated group feature emphasizing each target component is obtained by the element-wise product of $\bar{M}_k$ and $F_k$:
% By processing original ASC based part map with group convolution layer, the semantic information is richer and the size can be adjusted to the same as input feature while keeping different part of activation map independent. Second, in order to get ASC activated feature, we divide the input feature into k parts along the channel dimension and apply element-wise product between input feature and ASC based activation map. 
\begin{equation}
    P_k=\bar{M}_k \odot F_k
\end{equation}
% \begin{flalign}
% \label{deqn_ex1}
% &&P_i=\bar{M}_i\odot F_i    \qquad  i=0,1\cdots k-1
% \end{flalign}
% where $P_i$ is i-th part of ASC activated feature and $F_i$ is i-th part of input feature. Activated by ASC based activation map, the feature contains information of different components.

% 3) \emph{Physics informed attention}
\subsubsection{Physics Informed Attention}

The objective of the PIA module is to re-weight the importance of features based on the local semantics of targets (SSE). Additionally, it aims to facilitate the interaction of local semantics among channels, thereby enhancing their informativeness (PIR). 

The conventional global average pooling (GAP) layer in the SE module takes into consideration the entire spatial information of features, which will cause the background to dominate the GAP result in the SAR target image. As a consequence, we propose a shared selective SE (SSE) module to concentrate on the primary semantic regions. The squeezed feature is obtained by a selective average pooling (SAP) layer instead of GAP, as shown in Fig. \ref{fig:savg}, where a threshold is defined to determine whether to preserve in $P_k$ to calculate the average pooling value.

For each sub-group feature $P_k$ with $\frac{C}{K}$ channels, the selective mask of the $i$-th feature within $P_k$ is denoted as $Mask_k^{(i)}$. It is obtained based on a defined threshold $\rho$:

% In SAR target recognition, the importance of various components is supposed to vary depending on the class of target. Inspired by [], PIA is proposed to establish relations between parts. In the process of PIA, channel attention of each part of $\bar{F_i}$ is extracted by a shared selective SE(SSE) module. Because targets are typically tiny in SAR images and the slice is dominated by background, the target information will be diluted in global average pooling of SE, thereby diminishing the effect of channel attention. 
% The selective average pooling (SAvg), as shown in Fig. 6 is proposed. The threshold operation is adopted to select area with intense backscattering:
\begin{equation}  
Mask_k^{(i)} = \left\{
\begin{array}{rcl}
1,       &      & {P_k^{(i)} \geq \rho}\\
0,       &      & {P_k^{(i)} < \rho}
\end{array} \right. 
\end{equation}
% where $P_{ij}$ is the j-th channel of i-th part and $P^{mask}_{ij}$ is the corresponding mask. 
Then, the selective average pooling of each feature is realized by:
\begin{equation}
    P\mathrm{sap}_k^{(i)} = \frac{\sum_{H\times W} Mask_k^{(i)} \odot P_k^{(i)}}{\sum_{H\times W} Mask_k^{(i)}}
\end{equation}
% Formally, a statistic $P^{SAvg}_{ij}\in\mathbb{R}^{k \times \frac{C}{k}\times 1 \times 1}$ is generated by a mask based average pooling, such that the j-th element of i-th part is calculated by: 
% \begin{equation}
% \begin{aligned}
% \label{deqn_ex1}
% P^{SAvg}_{ij}=\frac{\sum_{i=1}^H\sum_{j=1}^W(P^{mask}_{ij}\odot P_{ij})}{\sum_{i=1}^H\sum_{j=1}^W(P^{mask}_{ij})}
% \end{aligned}
% \end{equation}
The $k$-th sub-group output of SSE is achieved by a MLP layer as same as Equation (\ref{equ:se}):
% The rest part of SSE is the same as SE and the i-th part of SSE output can be represented by:
\begin{equation}
\begin{aligned}
\label{deqn_ex1}
P\mathrm{attn}_{k}=\sigma (W_1\delta (W_0(P\mathrm{sap}_k))
\end{aligned}
\end{equation}
% where $P^{SAvg}_{i}$ is the i-th part of $P^{SAvg}$. 
% SSE module is used to squeeze the spatial information input of feature and obtain the channel weight of different parts. 
Further, in order to establish the interaction between different local semantics, we design the physical information recalibration (PIR) implemented by a Softmax layer along the axis of target components:
\begin{equation}
\begin{aligned}
\label{deqn_ex1}
\bar{P}\mathrm{att}^{i}_{1},\cdots,\bar{P}\mathrm{att}^{i}_{K}=Softmax(P\mathrm{att}^{i}_{1},\cdots,P\mathrm{att}^{i}_{K}) \\
i=1,2,\cdots,\frac{C}{K}
\end{aligned}
\end{equation}
where $\bar{P}\mathrm{att}^{i}_{k}$ is the $i$-th attention in the $k$-th sub-group. PIR re-calibrates channel weight in order to interact the local semantics of various target components across channels. Finally, the output features of the PASE module $P^{out}$ can be obtained by:
% part-specific channel weight is fused by a channel-wise multiplication:
\begin{equation}
\label{deqn_ex1}
P^{out}_k=P_k\otimes\bar{P}\mathrm{att}_{k}
\end{equation}
% where $Y_i\in\mathbb{R}^{\frac{C}{k}\times H \times W}$ is i-th part of output. After re-weighted by the recalibrated channel attention, ASC activated feature establish the interaction between different parts. Finally, the output of PASE is obtained by a concat operation:
\begin{equation}
\begin{aligned}
\label{deqn_ex1}
P^{out}=Concat(P^{out}_1, P^{out}_2, \cdots, P^{out}_{K})
\end{aligned}
\end{equation}
% $P^{out}$ is the output of PASE and the output of data driven block and physics driven block are fused to obtain the output of PIHA:
% \begin{equation}
% \begin{aligned}
% \label{deqn_ex1}
% I = P^{out}+D^{out}
% \end{aligned}
% \end{equation}
%%总结一下  补全提取方法   SE图  选择性池化图和讲解

The aforementioned PASE attention method incorporates the physical information of the target as a robust prior, which not only activates the feature map to extract local sub-group features with semantic awareness but also aids in generating channel weights associated with the target component.
% first integrate the physics and deep learning by means of attention.  We activate different parts of input feature in PAM and reweight different parts of feature which interact different components with each other. 
% Thus, the physics informed feature with component relevance is obtained. 
% At last, we fuse the visual feature and physics informed feature to get the physics Inspired Hybrid feature.
%从上文的分析可以得出，我们提出的PASE方法第一次通过注意力的方式将物理信息与深度学习相结合，我们的方法基于物理信息对视觉特征进行激活，同时我们提出的PIR模块对不同part的权重进行调整，将不同part之间进行相连。我们最终得到了一种具物理引导下的备部件关联性的特征。

%在分类时，对不同的目标而言，其不同的part在分类时的重要性应该有所区别。比如T-72坦克的炮筒是明显的特征，因此该结构应该得到加强。因此我们提出PIA解决该问题。计上一步的输出为R，受启发，我们对具有ASC-part 特异性的特征进行如下处理。1.首先对上一步的特征分别输入一个相同的SEWeight，将特征的空间信息压缩，该模块用与获取输入特征的SE通道注意力，通过将具有part特异性的特征输入，我们可以获得具有part特异性的通道注意力Z。。。，Z反映了各个part的通道信息。2.对于同一个目标，其不同part之间应该分配不同的权重，但是在前文的步骤中，基于part的运算始终独立，因此我们提出PIR模块，该模块如下

%通过part维度，该模块的不同权重进行softmax运算，对权重进行相对调整，同时使得不同part权重的建立联系。最后经过调整后的通道注意力与进行通道上的相乘

\begin{figure}[!t]
\centering
\includegraphics[width=3in]{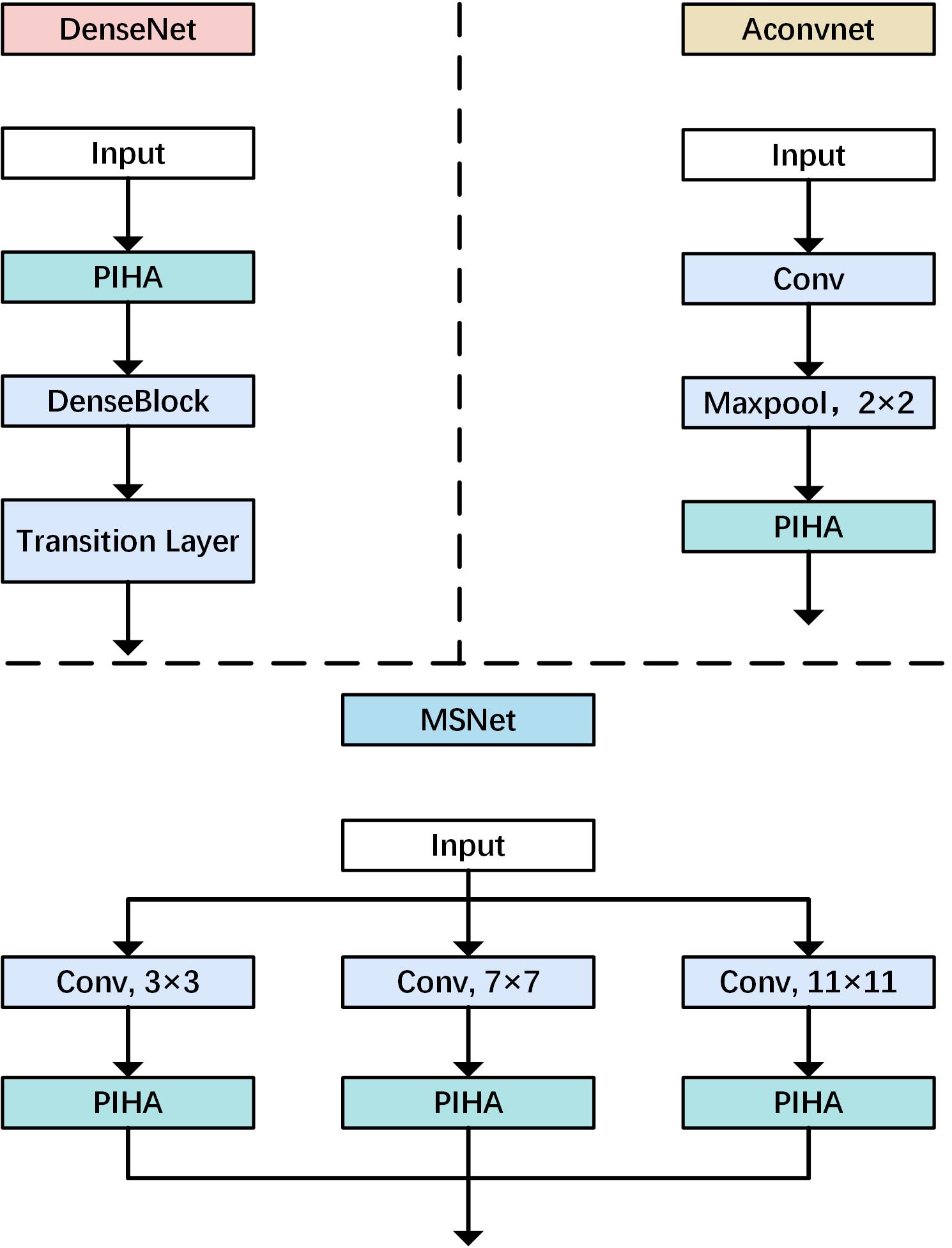}
\caption{Structure of PIHA enabled deep model.}
\label{fig:pihanet}
\end{figure}

\subsection{PIHA Enabled DNN}
\label{subsec:pihamodel}

The proposed PIHA can be embedded in different deep neural networks. In our work, three different backbones are explored, as illustrated in Fig. \ref{fig:pihanet}. Generally, PIHA is inserted after feature processing blocks such as convolution and max-pooling layers. For A-ConvNet \cite{7460942}, it is placed after every max-pooling layer. In DenseNet-121 \cite{8099726}, the proposed PIHA is followed by each DenseBlock and Transition Layer. In a multi-stream neural network such as MS-Net \cite{9780199}, PIHA is inserted after the convolution layer in each stream for the purpose of introducing physical information at different scales. By adaptively incorporating physics-aware semantics information, the PIHA-enabled deep neural network inherits the benefits of the original backbone and enhances its robustness and generalization capacity.

\section{Experiments}
\label{sec:exp}

\subsection{Experimental Settings}
%采集该数据集的传感器为高分辨率的聚束式合成孔径雷达，该雷达的分辨率为0.3m×0.3m。工作在X波段，所用的极化方式为HH极化方式。对采集到的数据进行前期处理，从中提取出像素大小为128×128包含各类目标的切片图像。该数据大多是静止车辆的SAR切片图像，包含多种车辆目标在各个方位角下获取到的目标图像。在该数据集中包含一个该计划推荐使用的训练集和测试集。训练集是雷达工作俯仰角为17时所得到的目标图像数据，包括3大类:BTR70（装甲运输车），BMP2（步兵战车），T72（坦克）；测试集是雷达工作俯仰角为时所得到的目标图像数据，该数据集也包含3大类，BMP2、T72、BTR70。各种类别的目标还具有不同的型号，同类但不同型号的目标在配备上有些差异，但总体散射特性相差不大。

\begin{table*}[!htbp]
\centering
\caption{The once-for-all (OFA) evaluation protocols for MSTAR dataset.}
\label{tab:data}
\resizebox{\linewidth}{!}{
\begin{tabular}{|c|c|c|c|c|c|c|c|c|c|c|c|c|}
    \hline
        & \multicolumn{3}{c|}{\textbf{Train/Val}} & \multicolumn{3}{c|}{\textbf{OFA-1}} & \multicolumn{3}{c|}{\textbf{OFA-2}} & \multicolumn{3}{c|}{\textbf{OFA-3}} \\
        \hline
        \textbf{Class} & \textbf{Serial} & \textbf{Dep.} & \textbf{No.} & \textbf{Serial} & \textbf{Dep.} & \textbf{No.} & \textbf{Serial} & \textbf{Dep.} & \textbf{No.} & \textbf{Serial} & \textbf{Dep.} & \textbf{No.} \\ 
        \hline
       \multirow{3}*{\textbf{BMP-2}} & 9563 & 17° & 233 & 9563 & 15° & 195 & 9563 & 15° & 195 & / & / & / \\ 
       \cline{2-13}
         & / & / & / & / & / & / & 9566 & 15° & 196 & / & / & /  \\ 
        \cline{2-13}
         & / & / & / & / & / & / & C21 & 15° & 196 & / & / &  / \\ 
        \hline
        \textbf{BTR-70} & C71 & 17° & 233 & C71 & 15° & 196 & C71 & 15° & 196 & / & / & /  \\ 
        \hline
        \multirow{3}*{\textbf{T-72}} & 132 & 17° & 232 & 132 & 15° & 196 & 132 & 15° & 196 & / & / & /  \\ 
        \cline{2-13}
         & / & / & / & / & / & / & 812 & 15° & 195 & / & / & /  \\ 
        \cline{2-13}
         & / & / & / & / & / & / & S7 & 15° & 191 & / & / &  / \\ 
        \hline
        \textbf{BTR-60} & k10yt7532 & 17° & 256 & k10yt7532 & 15° & 195 & k10yt7532 & 15° & 195 & / & / &  / \\ 
        \hline
        \textbf{2S1} & b01 & 17° & 299 & b01 & 15° & 274 & b01 & 15° & 274 & b01 & 15°/30°/45° & 274/288/303  \\ 
        \hline
        \textbf{BRDM-2} & E-71 & 17° & 298 & E-71 & 15° & 274 & E-71 & 15° & 274 & E-71 & 15°/30°/45° & 274/420/423  \\ 
        \hline
        \textbf{D7} & 92v13015 & 17° & 299 & 92v13015 & 15° & 274 & 92v13015 & 15° & 274 & / & / &  / \\ 
        \hline
        \textbf{T-62} & A51 & 17° & 299 & A51 & 15° & 273 & A51 & 15° & 273 & / & / &  / \\ 
        \hline
        \textbf{ZIL-131} & E12 & 17° & 299 & E12 & 15° & 274 & E12 & 15° & 274 & / & / & /  \\ 
        \hline
        \textbf{ZSU-234} & d08 & 17° & 299 & d08 & 15° & 274 & d08 & 15° & 274 & d08 & 15°/30°/45° & 274/406/422 \\
        \hline
        \textbf{Total} & \multicolumn{3}{c|}{2747} & \multicolumn{3}{c|}{2425} & \multicolumn{3}{c|}{3203} & \multicolumn{3}{c|}{3093} \\
    \hline
    \end{tabular}}
\end{table*}

\subsubsection{Once for All Evaluation Protocol} 

Our experiments are mainly conducted on the MSTAR dataset \cite{mstar} with the proposed once for all (OFA) evaluation protocol.

% Traditionally, algorithms of SAR ATR are validated under two typical SAR geometric acquisition conditions, i.e. SOC and EOC.
The current deep learning based approaches can achieve very high performance on the traditional evaluation protocol, i.e., standard operation condition (SOC) and extended operation condition (EOC). Fig. \ref{fig:intro} shows the proposed PIHA-Net and the A-ConvNet performs 99.38\% and 99.25\% on SOC, 98.73\% and 97.36\% on EOC (version variants) respectively. There is limited scope for improving the performance of an algorithm based on the traditional evaluation protocol, hence diminishing the significance of the assessment. To this end, we propose a more rigorous OFA evaluation process.

Once for all denotes that, once the model has been trained and validated with known data, it is evaluated on multiple test sets with varying data distributions. It differs from the conventional SOC and EOC evaluations, which train multiple models to fit various test sets. OFA can more effectively evaluate the robustness and generalization capability of an algorithm.

The MSTAR dataset is collected by Sandia National Laboratory's SAR sensor operating at X-band and HH polarization. It contains several types of static military vehicles with variants, consisting of a full range of orientation angles and different depression angles. As shown in Table \ref{tab:data}, the samples observed in the depression angle of 17$^\circ$ are regarded as known data for training and validation. To evaluate the algorithm in the abundant and limited training data cases respectively, we randomly select 90\%, 50\%, 30\%, and 10\% samples as training set and correspondingly using 10\%, 50\%, 50\%, and 50\% samples in the rest 17$^\circ$ depression angle data as validation set, respectively.

% The majority of images, which contain a full range of viewing angles, are 128$\times$128 pixels chips of various static military vehicle with 0.3m$\times$0.3m resolution.
 % 90$\%$, 50$\%$, 30$\%$, 10$\%$ 
% 17$^{\circ}$ depression angle data are selected as training set while correspondingly,  10$\%$, 50$\%$, 50$\%$, 50$\%$ in the rest 17$^{\circ}$ depression angle data are selected as validation set. 

% Meanwhile, the testing set are selected under three condition i.e., similar conditions, version variation and depression angle variation. The details of our setting are shown in Table \ref{tab:data}.

The proposed OFA evaluation protocol contains three test scenarios. OFA-1 indicates that the test set has a similar data distribution to the training and validation sets, with the same types and serial numbers and a similar depression angle of 15$^\circ$. OFA-2 introduces four additional variants in BMP-2 and T-72 compared to OFA-1, 3203 samples in total. OFA-2 evaluates the robustness of recognition algorithms to various variants. OFA-3 is a mixture of different depression angles containing 15$^\circ$, 30$^\circ$, and 45$^\circ$, where only three categories are included with 3093 samples in total. There is a significant distribution drift in OFA-3 compared with training and validation set that makes the task more challenging. Consequently, OFA-3 evaluates the generalization capability of recognition algorithm to different observation conditions.

% The largest difference between OFA and traditional method evaluation protocol is OFA trains once while testing on three conditions.

\begin{table*}
\centering
\caption{Ablation studies conducted on DenseNet121 and A-ConvNet backbones.}
\label{tab:abl}
\resizebox{\linewidth}{!}{
\begin{tabular}{c c c c c c c c c c c c c c c}
    \toprule
    \multirow{2}{*}{Backbone} & Data-Driven & \multicolumn{2}{c}{Physics-Driven} & \multicolumn{3}{c}{90}& \multicolumn{3}{c}{50} \\
    \cline{2-10}
    &\makebox[0.05\textwidth][c]{SE} & \makebox[0.05\textwidth][c]{PAM} & \makebox[0.05\textwidth][c]{PIR} & OFA-1 & OFA-2 & OFA-3 & OFA-1 & OFA-2 & OFA-3 \\
    \midrule

        \multirow{5}{*}{DenseNet121}&\XSolidBrush & \XSolidBrush & \XSolidBrush & 95.37±1.04 & 91.73±0.76 & 60.37±2.66 & 91.60±1.82 & 88.45±1.58 & 60.18±1.57 \\
        &\Checkmark & \XSolidBrush & \XSolidBrush & 95.87±1.50  & 93.20±1.46 & 61.55±2.33 & 92.54±1.21 & 88.97±1.20 & 60.84±1.83 \\
        &\XSolidBrush & \Checkmark & \Checkmark & 78.86±4.31 & 74.09±4.4 & 50.67±2.1 & 66.8±1.63 & 62.62±2.05 & 48.05±0.91 \\
        &\Checkmark & 0.25 & 0 & 96.87±0.83 & 93.27±1.15 & 60.99±3.17 & \textbf{95.53±0.38} & 91.93±0.38 & 58.35±3.77 \\
        
        & \Checkmark & \Checkmark & \Checkmark & \textbf{97.41±0.99} & \textbf{94.4±1.66} & \textbf{62.07±2.4} & 95.32±1.3 & \textbf{91.97±1.72} & \textbf{61.81±1.16} \\
        \midrule
        \multirow{5}{*}{A-ConvNet}&\XSolidBrush & \XSolidBrush & \XSolidBrush &86.95±5.69 & 84.51±6.06 & 57.16±3.07 & 92.48±5 & 88.88±4.82 & 62.8±2.31 \\
        & \Checkmark & \XSolidBrush & \XSolidBrush & 86.18±5.01 & 83.9±3.12 & 57.24±2.57 & 92.89±1.32 & 88.74±2.63 & 57.86±2.78 \\
        &\XSolidBrush & \Checkmark & \Checkmark & 90.96±3.84 & 87.48±2.76 & 57.28±0.84 & \textbf{93.71±4.91} & \textbf{89.94±5.04} & \textbf{63.87±4.45} \\

        &\Checkmark & 0.25 & 0 & 89.53±7.44 & 86.98±7.32 & 57.83±5.3 & 90.66±3.76 & 87.66±3.84 & 59.41±3.24 \\

        & \Checkmark & \Checkmark & \Checkmark & \textbf{94.96±2.84} & \textbf{93.02±3.24} & \textbf{57.91±1.8} & 93.13±3.24 & 89.06±3.54 & 59.11±2.89 \\
    \bottomrule
    \toprule
    \multirow{2}{*}{Backbone} & Data-Driven & \multicolumn{2}{c}{Physics-Driven} & \multicolumn{3}{c}{30}& \multicolumn{3}{c}{10} \\
    \cline{2-10}
    &\makebox[0.05\textwidth][c]{SE} & \makebox[0.05\textwidth][c]{PFM} & \makebox[0.05\textwidth][c]{PAM} & OFA-1 & OFA-2 & OFA-3 & OFA-1 & OFA-2 & OFA-3 \\
    \midrule

        \multirow{5}{*}{DenseNet121}&\XSolidBrush & \XSolidBrush & \XSolidBrush & 84.72±1.13 & 80.29±1.21 & 57.44±2.56 & 65.58±3.49 & 59.32±3.21 & \textbf{56.63±3.18}\\
        & \Checkmark & \XSolidBrush & \XSolidBrush & 84.54±1.08 & 80.52±0.93 & 57.00±1.58 & 64.98±3.01 & 58.87±3.36 & 54.38±2.06\\
        & \XSolidBrush & \Checkmark & \Checkmark & 55.12±2.68 & 51.98±2.54 & 41.94±1.68 & 41.58±0.82 & 38.96±0.35 & 35.69±1.36 \\
        & \Checkmark & 0.25 & 0 & 89.48±1.01 & 85.6±1.08 & 60.13±1.24 & 70.71±2.37 & 65.16±1.78 & 54.01±2.19 \\
        
        & \Checkmark & \Checkmark & \Checkmark & \textbf{90.2±1.43} & \textbf{85.98±1.73} & \textbf{61.91±1.45} & \textbf{71.97±1.97} & \textbf{65.89±2.08} & 53.87±1.85 \\
        \midrule
        \multirow{5}{*}{A-ConvNet}&\XSolidBrush & \XSolidBrush & \XSolidBrush & 87.65±2.39 & 83.04±3.32 & 58.63±1.73 & 72.02±1.7 & 64.76±2.27 & 52.71±2.28 \\
        & \Checkmark & \XSolidBrush & \XSolidBrush & 90.38±2.85 & 85.1±3.38 & \textbf{59.44±2.92} & 73.53±4.32 & 66.82±3.46 & \textbf{54.07±4.18} \\
        &\XSolidBrush & \Checkmark & \Checkmark &82.62±2.14 & 76.23±2.27 & 56.8±1.52 & 62.22±3.11 & 56.73±3.23 & 49.11±1.36 \\

        & \Checkmark & 0.25 & 0 & 90.94±2.56 & 85.76±2.24 & 59.27±2.65 & \textbf{76.17±2.36} & 68.9±3.41 & 50.38±2.34 \\

        & \Checkmark & \Checkmark & \Checkmark & \textbf{91.18±2.19} & \textbf{86.49±2.49} & 58.78±1.03 & 76.11±1.93 & \textbf{69.6±2.95} & 53.22±1.56 \\
    \bottomrule
\end{tabular}}
\end{table*}

\subsubsection{Implementation Details} 

The target image chips are center cropped to 128$\times$128. The threshold $\rho$ in SAP is set to 0.05. The network parameters are randomly initialized without pre-training and the data augmentation is deactivated. The learning rate is set to 5e-4, 5e-3, and 5e-4 for DenseNet121 \cite{8099726}, A-ConvNet \cite{7460942}, and MS-Net backbones, respectively. The batchsize is 32. Stochastic gradient descent (SGD) is applied with momentum rate of 0.9 and weight decay of 0.001. A strategy of early stopping is implemented, which entails ceasing training if the accuracy of the validation set does not improve within 200 epoch. The utmost number of training epochs is 1000. Each experiment is repeated five times in order to record the mean and standard deviation.

\subsection{Ablation Studies}

In this section, we report ablation study results based on DenseNet-121 and A-ConvNet backbones, two representative architectures of deep and shallow neural networks, to investigate the effectiveness of PIHA module.

\subsubsection{Data-driven SE}
The original SE block aims to recalibrate the feature importance among channels. The SE weights are determined by the global information of feature maps. Table \ref{tab:abl} reported the performance of the baseline backbone and the SE-Net counterpart in the first and second row. Noted that when sufficient training data is available, the deeper model outperforms the shallow model, whereas the shallow architecture performs better when training data is scarce. Interestingly, we can observe that the data-driven SE block improves the performance of deeper architecture with sufficient training data, and the shallow architecture with insufficient training data, more significantly than the other cases.

\subsubsection{Physics-aware SE}
When only physics-driven attention is enabled, as shown in the third row in Table \ref{tab:abl}, the performance of shallow and deeper architectures differs significantly. DenseNet-121 fails with only PASE module, whereas A-ConvNet improves when trained with 90\% and 50\% of the data, but fails with very few training examples. As a comparison, the combination of PASE and data-driven SE improves the performance. The phenomenon manifest the different working mechanism of data-driven and physics-aware attentions. PASE module provides strong knowledge prior of local semantics related to target components, and the trainable parameters in PASE enable the physical priors to be more adaptable and flexible during the optimization process. The very limited training samples, as well as the very deep layer-wise information process, will result in more strict constraint of physical priors with losing accuracy. Therefore, the combination of knowledge aware PASE and flexible data-driven SE collaborate to improve the performance of DenseNet-121 significantly.

\subsubsection{Physical information Recalibration}
The proposed physics informed attention in the designed PASE module includes: SSE to reweight the feature map within each target component, and PIR to establish the interaction among different components. Table \ref{tab:abl} reported the effectiveness of the proposed PIR by comparing the results in the fourth and fifth row. The parameters 0.25 and 0 indicate that the output of SSE is multiplied by 0.25 to acquire the final PASE attention without PIR for the purpose of comparing the performance of enabling PIR with Softmax under the condition of four target components. It can be inferred from the results that the interaction between components can further enhance the performance in most cases.

\subsection{Effectiveness of PIHA on Different Architectures}

To evaluate the effectiveness of the proposed PIHA module, we apply different backbones with PIHA modules as illustrated in Section \ref{subsec:pihamodel}. The results are given in Table \ref{tab:pihanet}, where DenseNet-121 \cite{8099726}, A-ConvNet \cite{7460942}, and MS-Net are explored. MS-Net is modified on the basis of MS-CVNets \cite{9780199} where we replace the complex-valued layers with real number structure and deepen the channel of feature. The specific definition of models can be found at \url{https://github.com/XAI4SAR/PIHA/model/}.

The results are recorded in Table \ref{tab:pihanet}, where the robustness and generalization ability of the algorithm are comprehensively evaluated on the condition of abundant and limited data. The PIHA enabled neural network can achieve better performance than the original one in most cases. The effectiveness of PIHA on different architectures can be proved.

% 在这里介绍 MS-Net 的结构？

\begin{table*}
\centering
% \scriptsize
\caption{The performance comparison of different backbones and their PIHA enabled versions.}
\label{tab:pihanet}
\begin{tabular}{c c c c c c c}
    \toprule
    \multirow{2}{*}{Model} & \multicolumn{3}{c}{90}& \multicolumn{3}{c}{50} \\
    \cline{2-7}
    & OFA-1 & OFA-2 & OFA-3 & OFA-1 & OFA-2 & OFA-3  \\
    \midrule

        DenseNet121 & 95.37±1.04 & 91.73±0.76 & 60.37±2.66 & 91.60±1.82 & 88.45±1.58 & 60.18±1.57 \\
        \rowcolor{gray!20} + PIHA & \textbf{97.41±0.99} & \textbf{94.4±1.66} & \textbf{62.07±2.4} & \textbf{95.32±1.3} & \textbf{91.97±1.72} & \textbf{61.81±1.16} \\
        \midrule
        
        A-ConvNet & 86.95±5.69 & 84.51±6.06 & 57.16±3.07 & 92.48±5 & 88.88±4.82 & \textbf{62.8±2.31} \\
        \rowcolor{gray!20} + PIHA & \textbf{94.96±2.84} & \textbf{93.02±3.24} & \textbf{57.91±1.8} & \textbf{93.13±3.24} & \textbf{89.06±3.54} & 59.11±2.89 \\
        \midrule
        
        MS-Net & 97.91±0.94 & 95.33±0.87 & 64.2±2.03 & 97.44±0.24 & \textbf{94.41±0.55} & 65.84±0.58 \\
        \rowcolor{gray!20} + PIHA & \textbf{98.22±0.38} & \textbf{96.22±0.65} & \textbf{66.1±3.09} & \textbf{97.59±0.23} & 94.36±0.54 & \textbf{66.47±1.35} \\
    \bottomrule
    \toprule
    \multirow{2}{*}{Backbone} & \multicolumn{3}{c}{30}& \multicolumn{3}{c}{10} \\
    \cline{2-7}
     & OFA-1 & OFA-2 & OFA-3 & OFA-1 & OFA-2 & OFA-3  \\
    \midrule

        DenseNet121 & 84.72±1.13 & 80.29±1.21 & 57.44±2.56 & 65.58±3.49 & 59.32±3.21 & \textbf{56.63±3.18}\\
        \rowcolor{gray!20} + PIHA & \textbf{90.2±1.43} & \textbf{85.98±1.73} & \textbf{61.91±1.45} & \textbf{71.97±1.97} & \textbf{65.89±2.08} & 53.87±1.85 \\
        \midrule
        
        A-ConvNet & 87.65±2.39 & 83.04±3.32 & 58.63±1.73 & 72.02±1.7 & 64.76±2.27 & 52.71±2.28\\
        \rowcolor{gray!20} + PIHA & \textbf{91.18±2.19} & \textbf{86.49±2.49} & \textbf{58.78±1.03} & \textbf{76.11±1.93} & \textbf{69.6±2.95} & \textbf{53.22±1.56} \\
        \midrule
        
        MS-Net & 92.94±0.73 & 88.33±0.82 & 63.9±2.35 & 78.43±0.73 & \textbf{72.81±1.01} & 59.82±0.77\\
        \rowcolor{gray!20} + PIHA & \textbf{93.41±0.83} & \textbf{89.31±0.9} & \textbf{64.88±1.22} & \textbf{78.56±0.74} & 72.49±0.37 & \textbf{60.23±0.98}\\
    \bottomrule
\end{tabular}
\end{table*}

\begin{table*}
\centering
\caption{The comparative study with some state-of-the-art methods. Note that * is our implementation, as there is no publicly available source code. All hybrid modeling methods apply the unified physical parameters of ASC model optimized by OMP algorithm. \textbf{Bold} and \underline{Underline} highlight the best and the second best result, respectively.}
\label{tab:comparison}
\resizebox{\linewidth}{!}{
\begin{tabular}{c c c c c c c c c}
\toprule
    \multirow{2}{*}{Method} & \multirow{2}{*}{\makecell[c]{Modeling}} & \multirow{2}{*}{\makecell[c]{Input}} & \multicolumn{3}{c}{90} & \multicolumn{3}{c}{50} \\
    \cline{4-9}
    & & & OFA-1 & OFA-2 & OFA-3 & OFA-1 & OFA-2 & OFA-3 \\

    \midrule
    ResNet18  & data-driven & amplitude & 90.22±6.98 & 89.5±6.92 & 52.28±5.71 & \underline{93.94±1.96} & \underline{93.05±2.22} & 56.82±1.73 \\

    A-ConvNet* & data-driven & amplitude & 86.95±5.69 & 84.51±6.06 & 57.16±3.07 & 92.48±5 & 88.88±4.82 & \underline{62.8±2.31} \\

    FEC*  & hybrid & complex & 92.32±4.41 & 86.22±5.03 & 58.76±5.18 & 86.23±5.62 & 81.34±5.83 & 55.03±2.35 \\

    ESF*& hybrid & amplitude & 89.81±5.08 & 85.37±6.03 & 57.98±1.88 & 91.68±3.53 & 88.61±3.03 & 54.04±4.53 \\

    CA-MCNN*  & hybrid & complex & 94.56±2 & 92.76±1.72 & 49.35±5.07 & 93.34±3.08 & 91.56±2.37 & 51.81±4.23 \\

    MS-CVNets & data-driven & complex & \underline{96.77±1.74} & \underline{94.34±1.22} & \textbf{66.83±3.96} & 93.43±1.89 & 90.78±1.83 & 62.56±3.71 \\

    \rowcolor{gray!20} MS-PIHA (Ours) & hybrid & flexible & \textbf{98.22±0.38} & \textbf{96.22±0.65} & \underline{66.1±3.09} & \textbf{97.59±0.23} & \textbf{94.36±0.54} & \textbf{66.47±1.35} \\
\bottomrule
\toprule
\multirow{2}{*}{Method} & \multirow{2}{*}{\makecell[c]{Modeling}} & \multirow{2}{*}{\makecell[c]{Input}} & \multicolumn{3}{c}{30} & \multicolumn{3}{c}{10} \\
    \cline{4-9}
    & & & OFA-1 & OFA-2 & OFA-3 & OFA-1 & OFA-2 & OFA-3  \\

    \midrule
    ResNet18 & data-driven & amplitude & \underline{92.08±3.25} & \textbf{90.23±3.34} & 50.4±1.05 & 71.67±6.19 & 67.37±4.44 & 40.78±5.8 \\

    A-ConvNet* & data-driven & amplitude & 87.65±2.39 & 83.04±3.32 & \underline{58.63±1.73} & 72.02±1.7 & 64.76±2.27 & \underline{52.71±2.28} \\

    FEC* & hybrid & complex & 68.43±7.72 & 64.48±6.16 & 51.12±1.81 & 57.84±3.47 & 54.04±3.91 & 43.44±5.97 \\

    ESF* & hybrid & amplitude & 89.38±1.47 & 85.26±1.79 & 56.82±1.43 & 75.92±4.79 & 71.31±3.96 & 50.21±5.69 \\

    CA-MCNN* & hybrid & complex & 90.99±1.11 & 86.6±1.5 & 52.72±2.94 & \textbf{80.02±1.48} & \textbf{73.19±1.08} & 44.18±1.21 \\

    MS-CVNets & data-driven & complex & 81.33±0.95 & 77.5±0.97 & 56.02±1.11 & 49.11±4.27 & 44.63±3.74 & 38.96±5.78 \\

    \rowcolor{gray!20} MS-PIHA (Ours) & hybrid & flexible & \textbf{93.41±0.83} & \underline{89.31±0.9} & \textbf{64.88±1.22} & \underline{78.56±0.74} & \underline{72.49±0.37} & \textbf{60.23±0.98}  \\
\bottomrule
\end{tabular}}
\end{table*}

\subsection{Comparative Study with SOTA Methods}

We compare the proposed PIHA enabled deep model based on MS-Net architecture (MS-PIHA) with other SAR target recognition approaches, including the data-driven real-valued neural network (ResNet-18 \cite{7780459}, A-ConvNet \cite{7460942}), complex-valued neural network (MS-CVNets \cite{9780199}), and the hybrid modeling approaches integrating ASC model (FEC \cite{zhangFECFeatureFusion2021}, ESF \cite{9896887}, CA-MCNN \cite{9883598}). To make the comparison fair enough, we apply the same ASC optimization results with our work and all experiments follow the proposed OFA evaluation protocols. The results are recorded in Table \ref{tab:comparison}, where * represents our implementations since there is no publicly available source code. The hyper-parameter setting follows the original papers. All models were trained without any data augmentation.

As shown in Table \ref{tab:comparison}, the OFA evaluation results of the comparative methods are not as good as those presented in the original papers evaluated on SOC and EOC. This indicates the proposed OFA evaluation is more rigorous and makes the task more challenging, allowing a more reliable assessment of generalization capability and robustness. The proposed MS-PIHA scores 8 first-place and 4 second-place results in terms of the mean value of the results among the twelve test scenarios, and achieves smaller fluctuation compared with the four first-place results. In the most challenging case of OFA-3 with 10\% training data, our method outperforms the second-best result by 7.52\%. It also outperforms other hybrid modeling methods with the same ASC parameters. The variability in performance among three hybrid modeling approaches (FEC, ESF, CA-MCNN) with the same ASC settings indicates that the outcomes of the algorithm are significantly influenced by the results of the physical model. As a comparison, the proposed PIHA enabled model leverages the semantic prior of physical parameters rather than their specific values, resulting in enhanced performance with increased adaptability and flexibility. 

Note that the data-driven method MS-CVNets, taking the complex-valued data as the input, achieve remarkable results with sufficient training data. With very limited training samples, i.e., only 10\% for training, the hybrid modeling methods, such as CA-MCNN and our PIHA enabled MS-Net, show superiority compared with data-driven approaches. 

In conclusion, hybrid modeling approaches that use physical information of SAR demonstrate a distinct advantage over data-driven methods when dealing with extremely limited labeled data. The physical information of SAR, such as physical parameters that describe target characteristics, provides strong knowledge prior. However, it exhibits limited flexibility and adaptability, since the model performance heavily depends on the quality of the physical information. The knowledge prior guided model design can be more flexible to leverage the physical information of high quality while mitigating the impact of irrelevant prior.

\subsection{Physical Information Discussion}

\begin{figure}[!t]
\centering
\includegraphics[width=8cm]{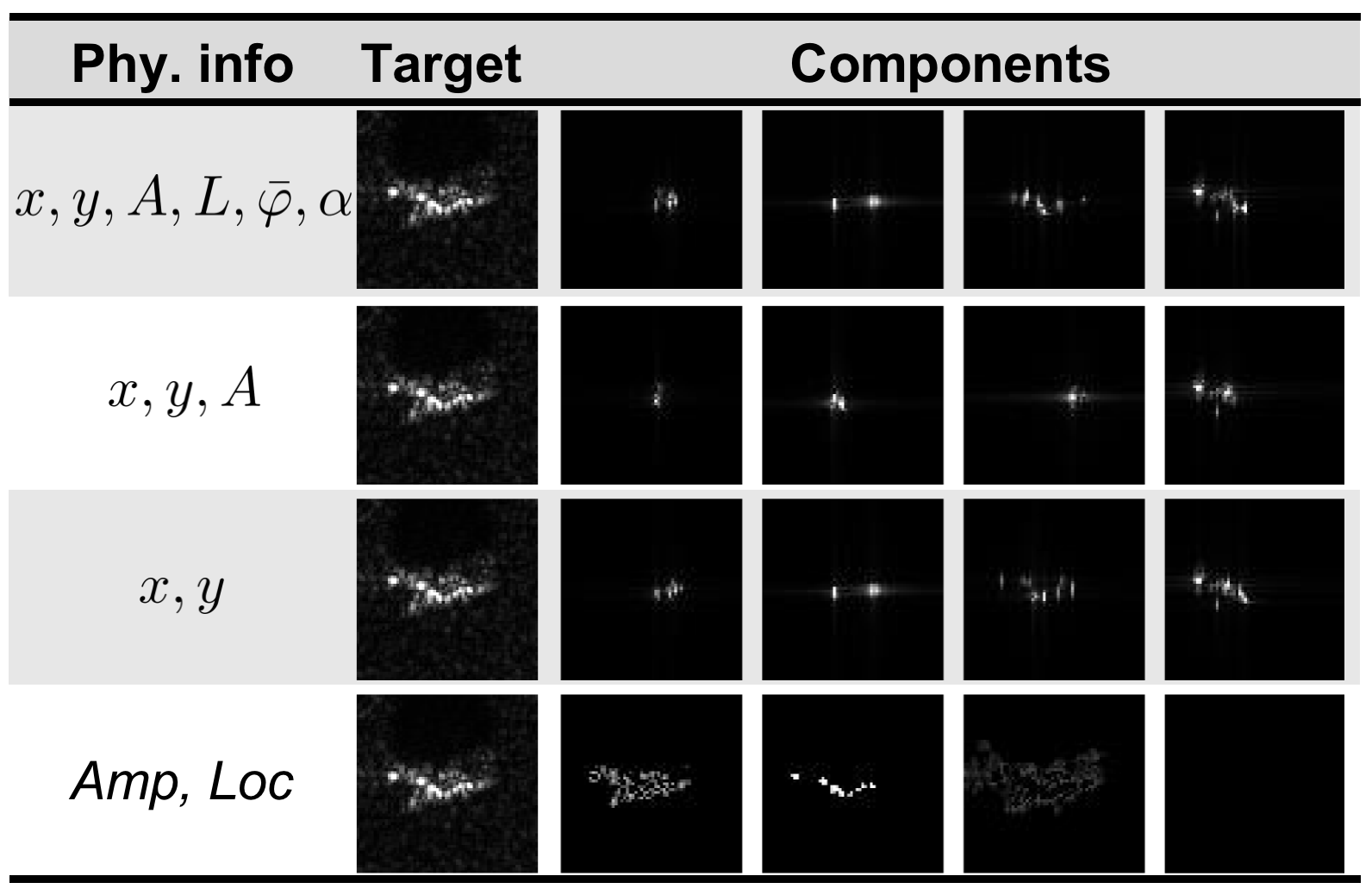}
\caption{The components clustering results based on different physical information, that are, $\{x,y,A,L,\bar{\varphi},\alpha\}$, $\{x,y,A\}$, $\{x,y\}$ of ASC parameters, and \{\textit{Amp, Loc}\} of the amplitude image, from top to bottom.}
\label{fig:exp_phyinfo}
\end{figure}

\begin{table*}
    \centering
\caption{The performance of PIHA module with various physical information.}
\label{tab:phy}
\begin{tabular}{c c c c c c c c}
\toprule
    \multirow{2}{*}{Model} &\multirow{2}{*}{\makecell[c]{Physics \\ Basis}}& \multicolumn{3}{c}{90}& \multicolumn{3}{c}{50} \\
    \cline{3-8}
    & & OFA-1 & OFA-2 & OFA-3 & OFA-1 & OFA-2 & OFA-3 \\

    \midrule
    DenseNet-121 & None & 95.37±1.04 & 91.73±0.76 & 60.37±2.66 & 91.60±1.82 & 88.45±1.58 & 60.18±1.57 \\
    \midrule
    +PIHA w/o ASC & Amp, Loc & 96.5±1.57 & 92.76±1.6 & \textbf{63.13±1.62} & 95.38±0.92 & 91.01±1.15 & 60.45±1.69 \\
    \midrule
    \multirow{3}{*}{+PIHA w/ ASC} & $x$,$y$ & 96.65±0.99 & 93.36±0.87 & 62.92±3.26 & 94.59±1.17 & 90.52±1.8 & 60.99±2.01 \\
    
    &$x$,$y$,$A$ & 96.87±1.37 & 93.51±1.5 & 62.81±2.08 & \textbf{95.48±0.79} & 91.68±0.41 & 60.89±2.22 \\

    &$x$,$y$,$L$,$\bar{\varphi}$,$\alpha$,$A$ & \textbf{97.41±0.99} & \textbf{94.4±1.66} & 62.07±2.4 & 95.32±1.3 & \textbf{91.97±1.72} & \textbf{61.81±1.16} \\
\bottomrule
\toprule
    \multirow{2}{*}{Model} &\multirow{2}{*}{\makecell[c]{Physics \\ Basis}} & \multicolumn{3}{c}{30} & \multicolumn{3}{c}{10} \\
    \cline{3-8}
    & & OFA-1 & OFA-2 & OFA-3 & OFA-1 & OFA-2 & OFA-3 \\

    \midrule
    DenseNet-121 & None & 84.72±1.13 & 80.29±1.21 & 57.44±2.56 & 65.58±3.49 & 59.32±3.21 & \textbf{56.63±3.18}\\
    \midrule
    +PIHA w/o ASC & Amp, Loc & 88.74±0.92 & 84.76±0.96 & 59.26±1.94 & 70.11±4.16 & 64.86±3.55 & 51.95±1.75 \\
    \midrule          
    \multirow{3}{*}{+PIHA w/ ASC} & $x$,$y$ & 90.15±0.95 & 85.94±1.79 & 60.78±4.04 & 70.75±2.36 & 65.18±2.03 & 54.51±2.07 \\
    
    &$x$,$y$,$A$ & 88.88±1.54 & 84.75±2.01 & 61.11±1.9 & 71.33±3.9 & 65.98±4.18 & 53.66±3.26 \\

    &$x$,$y$,$L$,$\bar{\varphi}$,$\alpha$,$A$ & \textbf{90.2±1.43} & \textbf{85.98±1.73} & \textbf{61.91±1.45} & \textbf{71.97±1.97} & \textbf{65.89±2.08} & 53.87±1.85 \\
\bottomrule
\end{tabular}
\end{table*}

In this section, we discuss the effectiveness of PIHA on the basis of different physical information. The ASC model, for example, contains several physical parameters to represent scattering centers. We generate the target components based on different combination of the physical parameters, such as the full description of attributed centers $\{x,y,A,L,\bar{\varphi},\alpha\}$, the location and complex amplitude $\{x,y,A\}$, and only the location $\{x,y\}$. As a comparison, we also attempt to use the backscattering intensity of the amplitude image, where the amplitude with each pixel is used for clustering to obtain the components. Fig. \ref{fig:exp_phyinfo} shows the different components clustering results based on various physical information of a SAR target.

The recognition performances of PIHA enabled DenseNet-121 based on different physical information are recorded in Table \ref{tab:phy}. Compared to the baseline model, the PIHA module enhances performance regardless of the physical information, even if only the amplitude information is applied without the ASC model. The full description of the scattering centers offers more sufficient physical information of SAR target, and thus the PIHA based on $\{x,y,A,L,\bar{\varphi},\alpha\}$ obtains better performance than the ones based on $\{x,y,A\}$ and $\{x,y\}$.

It demonstrates that the proposed PIHA can be formulated using diverse physical information, as long as it ensures the provision of local semantics for the target components. As a result, the PIHA enabled neural network can take flexible input without being restricted just to complex-valued data.

\begin{figure*}[!t]
\centering
\includegraphics[width=14cm]{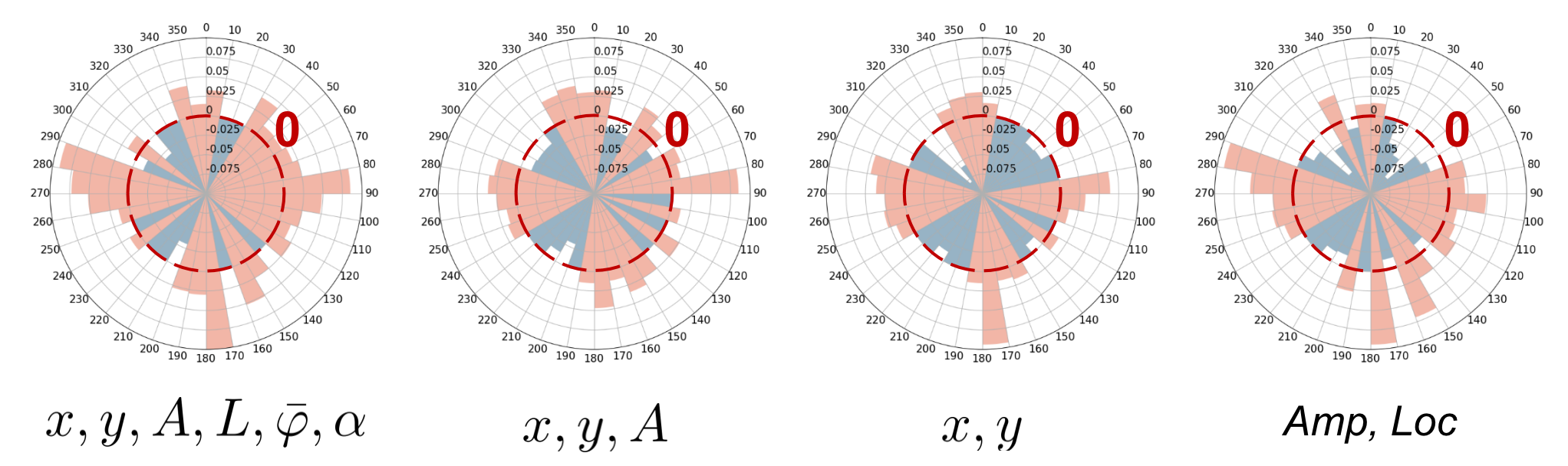}
\caption{The recognition rate difference between the proposed hybrid attention PIHA and the data-driven SE attention model, displayed in the Coxcomb Chart to represent the results in every 10$^\circ$ azimuth angle. The red(blue) sectors indicate how much PIHA(SE) outperforms SE(PIHA).}
\label{fig:exp_comparison}
\end{figure*}

\subsection{Attention Visualization}
\begin{figure*}[!t]
\centering
\includegraphics[width=17cm]{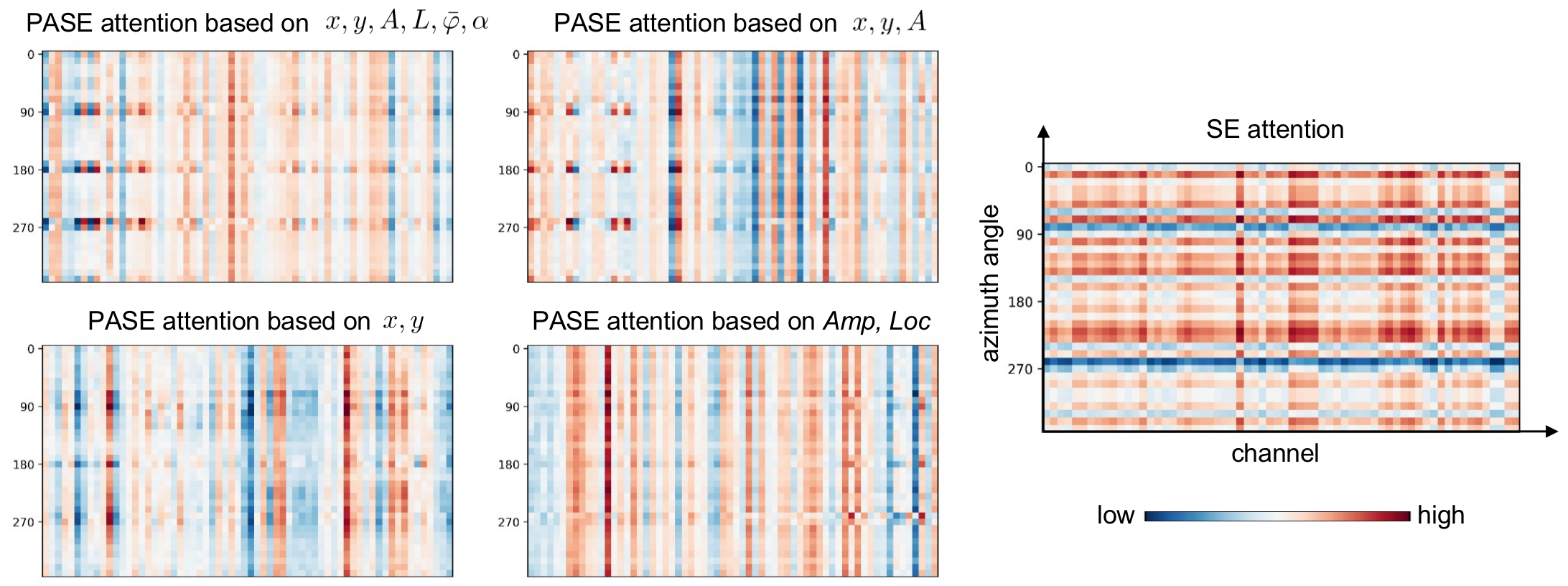}
\caption{The attention visualization of different PASEs compared with the data-driven SE. We calculate the averaged channel weights for all targets in every 10$^\circ$ azimuth angle interval. The x-axis represents channels and the y-axis represents azimuth angle of target.}
\label{fig:exp_attvis}
\end{figure*}

We calculate the averaged channel weights for all test images in every 10$^\circ$ azimuth angle interval, as shown in Fig. \ref{fig:exp_attvis}. The x-axis represents channels and the y-axis denotes azimuth angles. The colorbar demonstrates the higher/lower attention weights in red/blue. The channel weights of data-driven SE and PASE based on different physical information are given.

As we can referred in the Fig. \ref{fig:exp_attvis}, the data-driven SE attention is less discriminating cross channels. The channel attention conveys what to pay attention to, that is, the importance of each channel's feature is modeled. In the deep layers, where the spatial information is discarded to a certain extent, the channel information contains rich semantics. Observing from each row, for example, the SE attention weights are with homogeneous high values for those targets with azimuth angles near 220$^\circ$. It indicates that the data-driven SE is inclined to assign similar channel weights for each input SAR target image. As a comparison, the PASE attentions are more discriminative cross channels where each sub-group of channels represent a local semantic related to target component, especially for those targets near some specific azimuth angles such as 90$^\circ$, 180$^\circ$, and 270$^\circ$. Correspondingly, the recognition accuracies of PIHA-Net near these specific angles are prominently superior with SE-Net, as shown in Fig. \ref{fig:exp_comparison}. Additionally, PASE attention based on ASC parameters behaves remarkable distinctions along azimuth angles. It is well-known that SAR targets vary dramatically with azimuth angles, and the ASC model describes the target considering the changes of scattering center with azimuth looks. Consequently, the proposed physics-driven PASE is able to be aware of the physical parameters, and describes the semantic related channel weight more significantly. Notably, the PASE attention based on the full description of ASC model behaves much more discriminative than the ones based on less physical parameters. The (\textit{Amp, Loc}) based PASE attention is rarely discriminating cross azimuth angles due to the simple physical information, but it still remains differentiation cross channels better than SE attention.

\section{Conclusion}
\label{sec:conclusion}

An effective approach to incorporating domain expertise involves designing a model architecture that complies with established scientific principles. For SAR targets, the combination of local components provides physically explainable semantics that should benefit target recognition. To this end, we propose a physics inspired hybrid attention mechanism where the design consideration is informed with a comprehensive understanding of SAR target characteristics. In order to evaluate the generalization ability of the method more rigorously, we propose the once-for-all evaluation protocol where the model is trained once and tested on different test settings. The experiments investigate the effects of PIHA inserted into different deep architectures, and leveraging various types of physical information of SAR targets. In addition, the method is assessed with limited training samples and compared with the state-of-the-art approaches. We analyze the different working mechanisms between physics driven and data driven attentions in-depth. To contribute to the community, the physical information applied in this paper and the source code are available to public that ensures the reproducibility of our work and the fairness of comparing with other hybrid modeling approaches.

\section{Acknowledgment}

This work was supported by the National Natural Science Foundation of China under Grant 62101459, U20B2068, 62293543, and the China Postdoctoral Science Foundation under Grant BX2021248.

\bibliographystyle{cas-model2-names}

\bibliography{cas-dc-template_final}



\end{document}